\documentclass[review,authoryear]{elsarticle}
\usepackage{array}
\newcolumntype{P}[1]{>{\centering\arraybackslash}p{#1}}
\usepackage{lineno,hyperref}
\usepackage{graphicx}  
\usepackage[labelsep=quad,indention=10pt]{subfig}
\usepackage{caption}
\usepackage{ dsfont }
\usepackage{float}
\usepackage{soul}
\usepackage{multirow}
\usepackage[dvipsnames]{xcolor}

\usepackage[english]{babel}
\usepackage{blindtext}

\usepackage{amsmath}
\usepackage{graphicx}
\graphicspath{ {./Figures/} }
\usepackage{caption}
\usepackage{graphicx}
\usepackage[ruled,vlined]{algorithm2e}

\modulolinenumbers[5]

\usepackage{todonotes}

\makeatletter
\def\fixedlabel#1#2{%
  \@bsphack%
  \protected@write\@auxout{}%
         {\string\newlabel{#1}{{#2}{\thepage}}}%
  \@esphack}
\makeatother

\journal{Computers and Electronics in Agriculture}

\begin{document}

\begin{frontmatter}

\title{Minimize Labeling Effort for Tree Skeleton Segmentation using an Automated Iterative Training Methodology}

\author{Keenan Granland}
\author{Rhys Newbury}
\author{Zijue Chen} 
\author{David Ting}
\author{Chao Chen\corref{mycorrespondingauthor}}

\cortext[mycorrespondingauthor]{Corresponding author}
\ead{chao.chen@monash.edu}

\address{Laboratory of Motion Generation and Analysis, Faculty of Engineering, Monash University, Clayton, VIC 3800, Australia}

\begin{abstract}
Training of convolutional neural networks for semantic segmentation requires accurate pixel-wise labeling which requires large amounts of human effort. The human-in-the-loop method reduces labeling effort; however, it requires human intervention for each image. This paper describes a general iterative training methodology for semantic segmentation, Automating-the-Loop. This aims to replicate the manual adjustments of the human-in-the-loop method with an automated process, hence, drastically reducing labeling effort. Using the application of detecting partially occluded apple tree segmentation, we compare manually labeled annotations, self-training, human-in-the-loop, and Automating-the-Loop methods in both the quality of the trained convolutional neural networks, and the effort needed to create them. The convolutional neural network (U-Net) performance is analyzed using traditional metrics and a new metric, Complete Grid Scan, which promotes connectivity and low noise. It is shown that in our application, the new Automating-the-Loop method greatly reduces the labeling effort while producing comparable performance to both human-in-the-loop and complete manual labeling methods.  
\end{abstract}

\begin{keyword}
Self-Training, Semantic Segmentation, Semi-supervised Learning, Computer Vision, Agricultural Engineering
\end{keyword}

\end{frontmatter}
\linenumbers

\section{Introduction}

Automation in agriculture is becoming increasingly viable as technology advances become more accessible. However, this technology is faced with many challenges due to the environmental variations on farms. Current robotic solutions require highly structured environments with flat ground, wide rows, and two-dimensional trellised tree structures. In reality, each farm has a unique method of shaping trees using different patterns depending on the apple variety, land, environment, and the farmers' preference.      

One of the significant challenges in fruit harvesting is modeling the complex environment. Accurately detecting tree skeletons can lead to increased total harvest and minimize damage to trees. Tree skeletons are typically occluded by leaves, fruits, and other farming structures. To mitigate occlusions, tree skeletons can be detected during dormant seasons \citep{Dorm}.          

\begin{figure}[!htbp]
    \centering
    \includegraphics[width=1\textwidth]{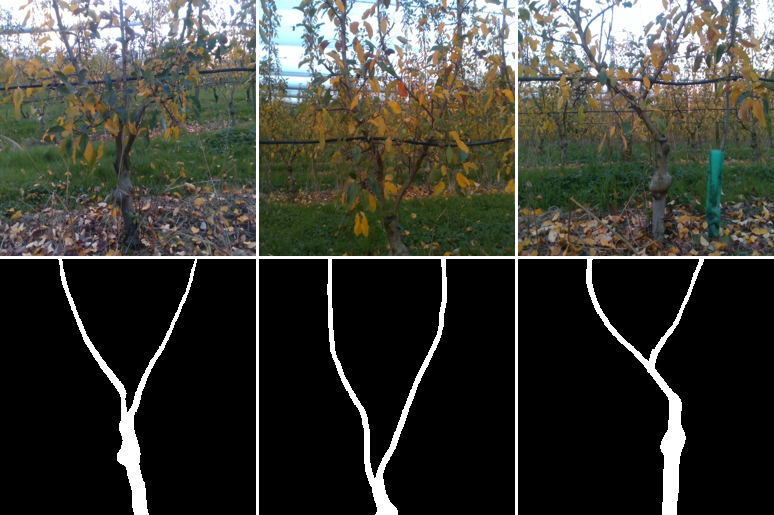}
    \caption{Examples of RGB image of a tree (top row) and the corresponding manually labeled data (bottom row).}
    \label{fig:IntroEg}
\end{figure}

In the process of detecting branches and trunks of trees, labeling is commonly done through object detection (box labels)~\citep{KANG2020105302}. However, for some applications, such as harvesting, thinning, and pruning of fruit trees, accurate modeling of the branches is required to proceed. As such, pixel-based labeling is required, which requires a large amount of human effort~\citep{chen2020semantic}, as shown in Figure \ref{fig:IntroEg}.  

To train deep learning models for semantic segmentation, accurate pixel by pixel labeling of possibly hundreds of thousands of images is needed for training~\citep{ulku2020survey}. To reduce this effort, researchers have explored Human-Machine Collaboration to reduce the human effort required in labeling data. For example, human-in-the-loop (HITL) assisted labeling, where a human annotator corrects the output of a neural network, reducing the total labeling effort. \citet{castrejon2017annotating} developed Polygon-RNN, which uses a polygon annotation tool to allow the human to quickly modify segmentation outputs speeding up the annotation by a factor $4.7$ across all classes in Cityscapes~\citep{Cordts2016Cityscapes}.

Recently, active learning has been used to automatically select a subset of data to be manually annotated. \citet{ravanbakhsh2019uncertaintydriven} use the discriminator of a generative adversarial network (GAN) to identify unreliable labels for which the export annotation is required. If the discriminator is confident, they use the generator to directly synthesize samples. \citet{pool-sampling-1} and \citet{pool-sampling-2} utilize attribute probabilities and effective relevance feedback algorithms to extract the most informative images for manual annotation from a large static pool. While these methods reduce the labeling effort required, they still require human assistance throughout the whole process.

In domains where additional labeling of images is not a viable option, other forms of semi-supervised learning have been proposed without the need for additional labeling. Semi-supervised learning is a general field in machine learning that trains supervised models using unlabeled data~\citep{semi-supervised}. The goal of semi-supervised learning is to achieve better results than what can be achieved with supervised learning when there is access to a larger amount of unlabeled data. To achieve this, many semi-supervised methodologies rely on a strong high-quality supervised baseline to build upon~\citep{semi-supervised}. The simplest form of semi-supervised learning is to utilize predictions from an unlabeled data-set in training directly~\citep{Lee2017}. However, this has issues with self-biasing as well as perpetuating mistakes. Many semi-supervised learning models attempt to resolve the perpetuating error problem by adding confident predictions (psuedo-labels) in to an updated training set. This is often called self-training~\citep{Lee2017}. For example, classifier models can output a confidence rating which can be used to select predictions for training based on a confidence threshold~\citep{semi-supervised} or custom scoring metrics external to the model can be used~\citep{cross-consistency}. Other methods have utilized these scores not to threshold but instead to weight the training labels created from the model~\citep{pseudo-labelling}. 

~\cite{IterativeLearning} proposed a dynamic curriculum learning strategy, which makes used of a self-training method to learn from feeding training images of increasing difficulty, which requires accurate difficulty evaluation. These methods have shown that the creation of accurate and specific scoring algorithms can be a powerful tool in enabling the use of unlabeled data for training. ~\cite{BNBS} proposed the Boundary-Aware Semi-Supervised Semantic Segmentation Network for high resolution remote sensing, that improves results without additional annotation workload by using custom processes to alleviate boundary blur. 

In this paper, we introduce a general iterative training method for Convolutional Neural Networks (CNNs) to increase the accuracy of the network. This work aims to train a CNN to accurately detect and label tree skeletons from a small set of data, minimizing the total human effort required. We trained a semantic segmentation CNN (U-Net) in an iterative process to provide pixel-wise labels for Y-shaped apple tree skeletons.  This was achieved using an automated method aiming to replicate the human adjustments from the human-in-the-loop process. The performance of each stage in the iterative training process is analyzed. Four different methods are outlined, explored, and compared to determine the effectiveness of each. Mean IOU and Boundary F1 are used to evaluate performance. Also, a custom scoring metric, Complete Grid Scan, is introduced to give a better comparison of the performance in this application.

The contributions of this paper are:
\begin{enumerate}
    \item Introduction of the Iterative Training Methodology for semi-supervised semantic segmentation learning.
    \item Introduction of a new metric, Complete Grid Scan.  
    \item Application of Automating-the-Loop to improve the generation of skeleton masks from a small dataset.
\end{enumerate}

The outline of the paper is as follows. We first introduce iterative training methods for CNNs in Section \ref{Methodology}. Implementation of these methods for the application of occluded apple tree detection is discussed in Section \ref{Implementation}. Performance is then evaluated and discussed in Section \ref{Results} before concluding in Section \ref{Conculsion} where future work is also outlined. 

\section{Iterative Training Methodology for Semantic Segmentation} \label{Methodology}

\begin{figure}[!htbp]
    \centering
    \includegraphics[width=1\textwidth]{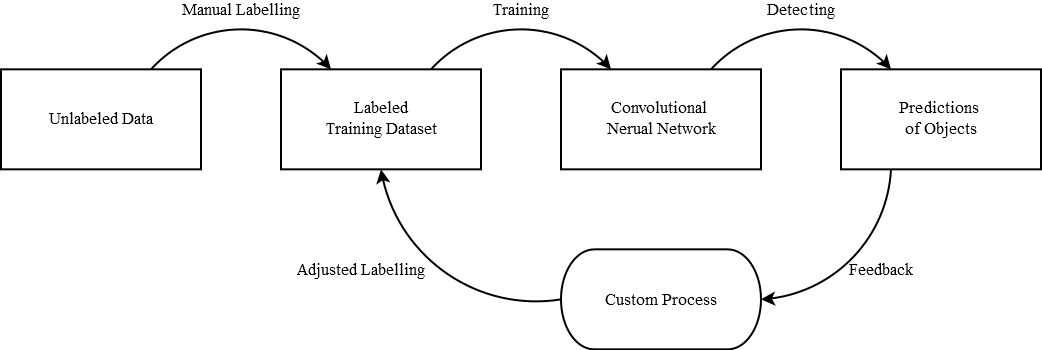}
    \caption{Flowchart outlining the Iterative Training Methodology.}
    \label{fig:SimpleFlowChart}
\end{figure}

We present a generalized iterative training loop for semantic segmentation in Figure \ref{fig:SimpleFlowChart} and Algorithm \ref{AL:ITM} describes the iterative process. It was observed that evaluating each pixel's confidence score does not reliably remove unwanted noises and false detections from predictions for self-training semantic segmentation. Therefore, we evaluate an entire thresholded prediction during the custom process. The custom process can range from a simple evaluation of the predictions to high-level repairing algorithms specific to the application. 

In semantic segmentation, each pixel is individually assessed to detect a whole object, as such, accuracy is highly valued. Contrasting to this, in the domain of object detection, where bounding box labels are used, evaluating and adjusting box labels based on confidence scores is more reliable, and pixel-level accuracy is not highly valued.

\begin{algorithm}[H]
\SetAlgoLined
\KwResult{CNN$_N$ and generated labels on unlabeled data}
 Manually label $S_{T0}$\;
 Remove $S_{T0}$ from $S_{U0}$\;
 Train CNN$_0$ on $S_{T0}$\;
 $i=1$\;
 \While{$|S_{U}|>0$}{
 \eIf{$i=1$}{
  Use CNN$_{i-1}$ to label $S_{Ui}$, generating $S_{Pi}$ and removing $S_{Ui}$ from $S_{U}$\;
  }{
  Use CNN$_{i-1}$ to label $S_{Ui}$ and $S_{Ai-1}$, generating $S_{Pi}$ and removing $S_{Ui}$ from $S_{U}$\;
  }
  Use the custom process on $S_{Pi}$, generating $S_{Ai}$ and $S_{Ri}$\;
  Create $S_{Ti}$ by combining $S_{T0}$ and $\displaystyle\sum_{k=1}^{i} S_{Ak}$\;
  Train CNN$_i$ using $S_{Ti}$\;
  $i = i + 1$
}
 \caption{Iterative Training Method}
  \label{AL:ITM}
\end{algorithm}

$S_{Ti}$ refers to the training set at iteration $i$. and $S_{T0}$ refers to the initial training set, an initial manually labeled dataset. To minimze human effort in labeling this should be as small as possible. $S_{Pi}$ is the  predictions, $S_{Ui}$ is the unlabeled datasets, $S_{Ai}$ is the pseudo-labeled images.

 The total unlabeled dataset is $S_{U}$ and is initially equal to $S_{T0} + \sum S_{Ui}$. $S_{U}$ reduces every iteration and after the final iteration $n$, $S_{U} = 0$. $S_{Ri}$ refers to the unsuccessfully labeled dataset at each iteration. The absolute values refer to the size of the dataset; for example, $|S_{U}|$ refers to the number of the unlabeled data.

There is no restriction on the size of the variable datasets ($S_{T0}$, $S_U$ and $S_{Ui}$) between iterations. Optimising these variables and selecting the custom process poses an interesting problem of future research to minimze human effort and maximising performance. 

The iterative approach aims to improve the CNN every iteration by increasing the training set size. Compared to using the custom process as only a post-processing filter for poorly labeled images, the automated iterative process aims to increase the number of high-quality labels in the training set. All unlabeled data will pass through the improved CNN every loop, allowing multiple chances to be accepted by the custom process.

\subsection{Confident Self-Training}
The confident self-training loop (CST) retrains the CNN iteratively by directly adding a set of predictions made from unlabeled data to the training set each iteration. There is no custom process (Figure \ref{fig:SimpleFlowChart}). This method is explored as a baseline for all other iterative methods. For CST, the custom process from Figure \ref{fig:SimpleFlowChart} is removed and the pseudo-code is represented in Algorithm \ref{AL:STL}.

\begin{algorithm}[!htbp]
\SetAlgoLined
\KwResult{CNN$_N$ and Iteratively Generated Labels}
 Manually label $S_{T0}$ 
 Remove $S_{T0}$ from $S_{U}$\;
 Train CNN$_0$ on $S_{T0}$\;
 $i=1$\;
 \While{$|S_{U}|>0$}{
  Use CNN$_{i-1}$ to label $S_{Ui}$, generating $S_{Pi}$ and removing $S_{Ui}$ from $S_U$\;
  Create $S_{Ti}$ by combining $S_{T0}$ and $\displaystyle\sum_{k=1}^{i} S_{Pk}$\;
  Train CNN$_i$ using $S_{Ti}$\;
  $i = i + 1$
}
 \caption{Confident Self-Training Loop}
 \label{AL:STL}
\end{algorithm}

\subsection{Filter-based Self-Training}
A filter-based self-training (FBST) learning system was explored as one of the baseline methods for this study. The filter provides a confidence of labels being correct. 

In this case, the custom process applied in both Figure \ref{fig:SimpleFlowChart} and Algorithm \ref{AL:ITM} is a filter, passing or rejecting predictions, allowing for fast training and labeling. In addition, the filter removes unwanted noise and false detections, generating pseudo-labels.

\subsection{Human-in-the-loop}
The HITL method reduces the human effort to create labeled images while producing a convolutional neural network (CNN) with comparable performance to a CNN trained on fully manually labeled images. The custom process applied in both Figure \ref{fig:SimpleFlowChart} and Algorithm \ref{AL:ITM}, involves an annotator adjusting the prediction dataset. Compared to manually labeling all images, it is estimated that HITL reduces labeling effort by a factor of 4.7~\citep{castrejon2017annotating}.

\subsection{Automating-the-Loop}
We introduce the Automating-the-Loop (ATL) method as an iterative method that uses a high-level automatic custom process on binary predictions from the CNN.
The process applied in both Figure \ref{fig:SimpleFlowChart} and Algorithm \ref{AL:ITM}, aims to replicate the human adjustments with an automatic process. This differs from traditional semi-supervised methods by not only removing labels after evaluation but adjusting and creating additional binary labels using features specific to the application. This can be achieved with various methods and may involve many stages of filtering, adjustment, addition and evaluation. 

Several different post-processing tools exist that can be applied in the process. As this is not expected to be completed in real-time, the quality of the automatic process should be the main focus as opposed to the computing time and method.

\section{Implementation} \label{Implementation}
We present an example of the methods described in the Section \ref{Methodology} for segmentation of partially occluded Y-shaped apple trees.

The unlabeled dataset contained $517$ RGB-D apple tree images taken from a commercial apple orchard in Victoria, Australia. The apple species is Ruby Pink, and the tree structure is 2D-open-V (Tatura Trellis). The apple trees are planted east to west, with $4.125$ m between each row and $0.6$ m between each tree in the row.  The images were taken by Intel RealSense D435, approximately 2m away from the tree canopies and around $1.3$ m above the ground. The original resolution of the images is 640 by 480 pixels.  The images were center cropped to $480$ by $480$ pixels and then rescaled to $256$ by $256$ pixels. All manual labels were created by labeling the images before cropping and scaling. 

\cite{purkait2019seeing} focused on masking the object that exists even behind the occlusion, a term often called `hallucinating' as it involves higher-level thinking to predict the mask based on the object as a whole. 

\subsection{Convolutional Neural Network Model}

\begin{figure}[!htbp]
    \centering
    \includegraphics[width=0.95\textwidth]{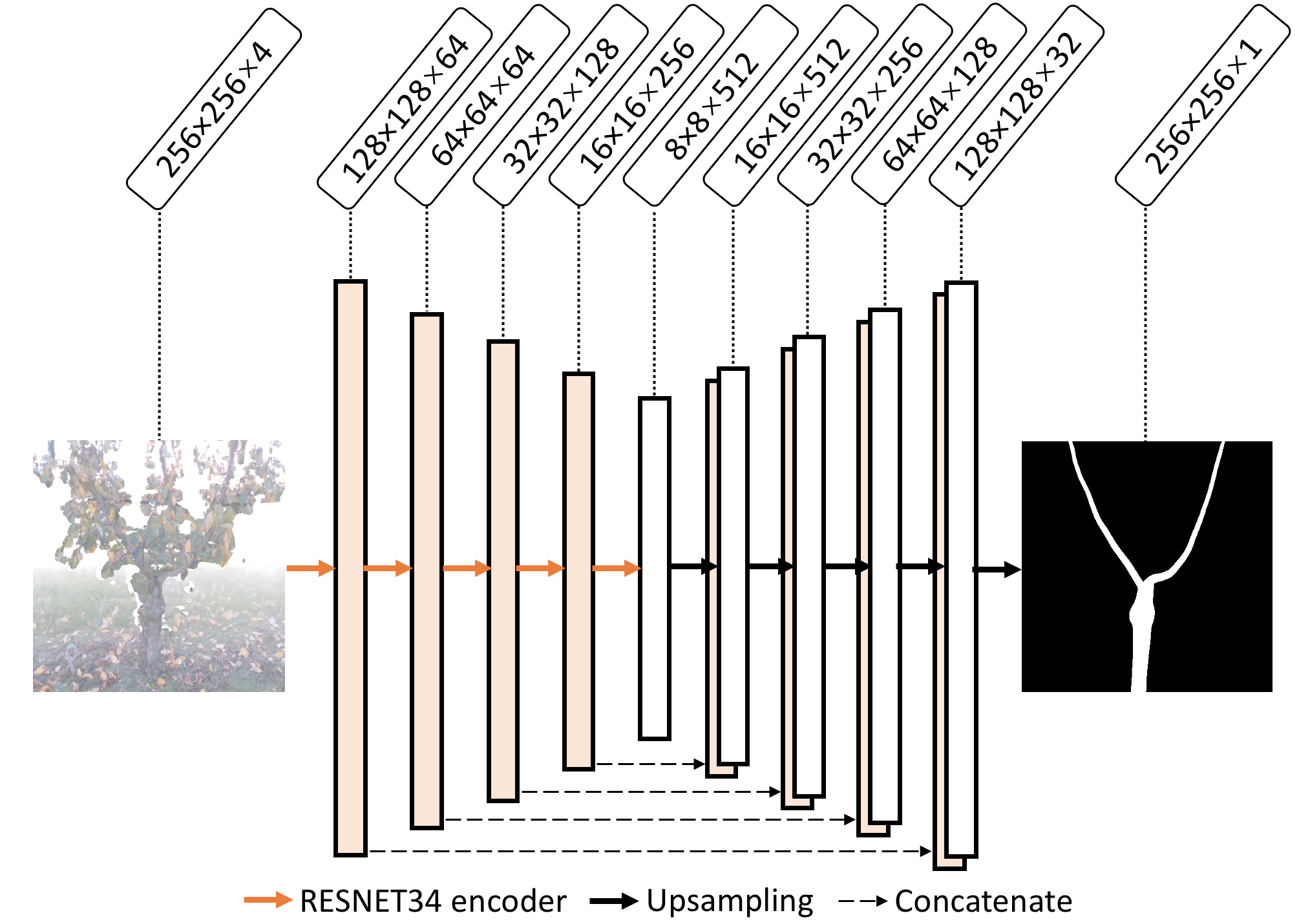}
    \caption{U-Net architecture}
    \label{fig:Unet structure}
\end{figure}

The neural network structure we used in this project was a U-Net~\citep{unet} which has an encoder-decoder structure using ResNet-34~\citep{resnet34} as the encoder (Figure \ref{fig:Unet structure},). This neural network configuration has been used extensively in pixel-wise semantic segmentation problems of irregularly shaped objects~\citep{Resnet-Unet, chen2020semantic} as well as occluded segmentation~\citep{purkait2019seeing}. The input to the network was an RGB-D image, which generated a binary mask output. This neural network configuration was trained using Tensorflow on a GTX1060 using the Adam optimizer and weighted dice loss~\citep{weighted_dice_loss}. The images were batched into batch sizes of 8 for training.

\subsection{Human-in-the-loop}
The HITL method was applied to train the U-Net~\citep{unet}. 

\begin{figure}[!htbp]
    \centering
    \includegraphics[width=0.95\textwidth]{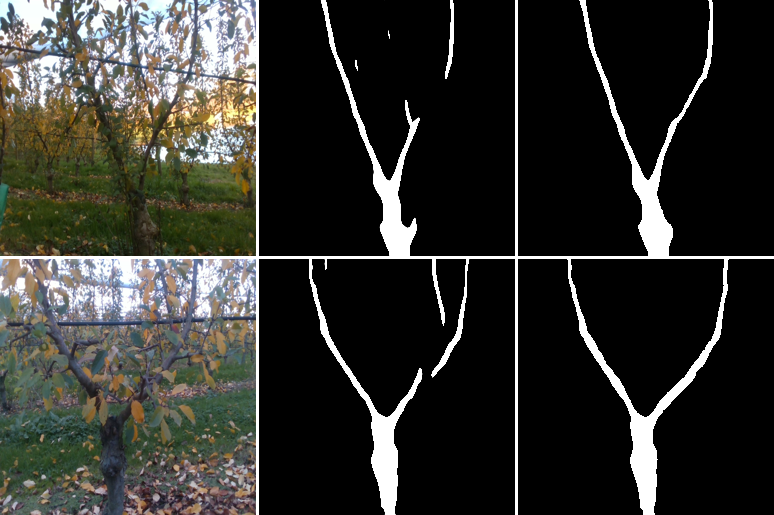}
    \caption{Examples of the HITL adjustment process. From left to right, RGB Image, CNN Prediction and Human Adjusted Label. Each row represents a unique image.}
    \label{fig:HITL_example}
\end{figure}

In this application, the main adjustments of the images were filling in small gaps, removing noise, and adjusting the thickness of some sections of the trees while using the RGB image as a reference (Figure \ref{fig:HITL_example}).

To adjust the predictions from the model, we used a similar tool to~\cite{castrejon2017annotating}. Firstly all vertices were converted into polygon points and then simplified using the Douglas-Peucker polygon simplification method~\citep{douglas-peucker}. These polygon points were then converted into the COCO format~\citep{COCO} and then adjusted in Intel's CVAT open-source annotation tool~\citep{cvat}.

\subsection{Filter-Based Self-Training}
\begin{figure}[!htbp]
    \centering
    \includegraphics[width=0.95\textwidth]{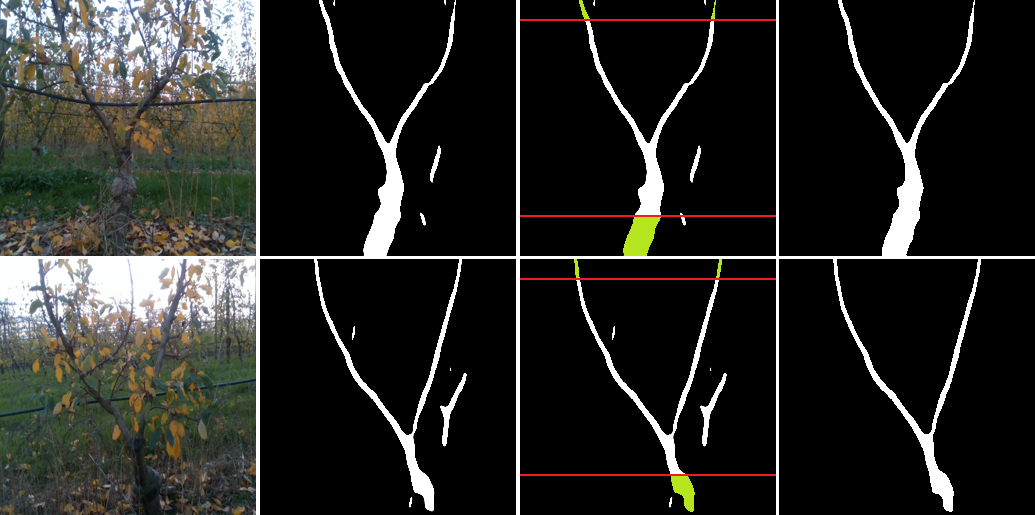}
    \caption{Examples of the FBST filtering process. From left to right, RGB Image, CNN Prediction , Y-shaped Tree Filter and adjusted pseudo-label. The Y-shaped Tree Filter indicates approved conditions in green, where the corresponding blob is then selected for the adjusted pseudo-label.}
    \label{fig:SemiSupervised_example}
\end{figure}

For FBST, we applied a blob filter to the predicted images. We defined a complete Y-shaped tree's minimum requirements: a blob must have at least two sections in the upper segment (top 20 pixel rows) and at least one section in the lower segment (bottom 40 pixel rows). We kept the largest blob meeting these requirements while all other blobs were removed. An example of this process is shown in Figure \ref{fig:SemiSupervised_example}. The complete pool of available unlabeled data was used in the first unlabeled dataset ($S_{U1}$) to allow the largest number of images to pass through the filter.

\subsection{Automating-the-Loop}
We developed an algorithm that involves several stages, one of which used an optimization tool, Genetic Algorithm (GA), to assist in repairing the prediction dataset. The automatic process occurred in three stages and aimed to replicate a similar outcome to a human manually adjusting the predicted images. These stages in order were: filtering, tree fitting (GA), and repairing.

\subsubsection{Filtering}
The filtering process aimed to remove unwanted noises and false detections from the image before the tree fitting process. The output of this stage was a filtered prediction (Figure \ref{fig:Automated_example} (3)). The filters used are as follows.
\begin{enumerate}
    \item \textbf{Small Blob Removal}: removes unwanted noise. For our image size of 256x256, small blobs less than $50$ pixels are removed. 
    \item \textbf{Different Tree Detection}: removes detection of other trees. By scanning the bottom rows of the image the initial center trunk position ($t_{pos}$) was found, any blobs with an center $x$ position  (horizontal axis) not within the range of $\pm 100$ pixels are removed.  
    \item \textbf{False Branch Detection}: removes false branch ends from the top of the image. Blobs with a center $y$ position $>240$ (top of image) and a height of $>5$ are removed.
    \item \textbf{False Trunk Detection}: removes false trunks or wooden poles detected. Blobs with a center $x$ position not within the range of $\pm 30$ of $t_{pos}$, with a height of $>80$ and a width of $<15$ are removed.
\end{enumerate}

\subsubsection{Fitting}
The filtering process outputs (filtered predictions) were then given as inputs to the tree fitting process. This process aimed to fit a tree template, aiming to generate the missing parts of the tree.

\begin{figure}[!htbp]
    \centering
    \includegraphics[width=0.9\textwidth]{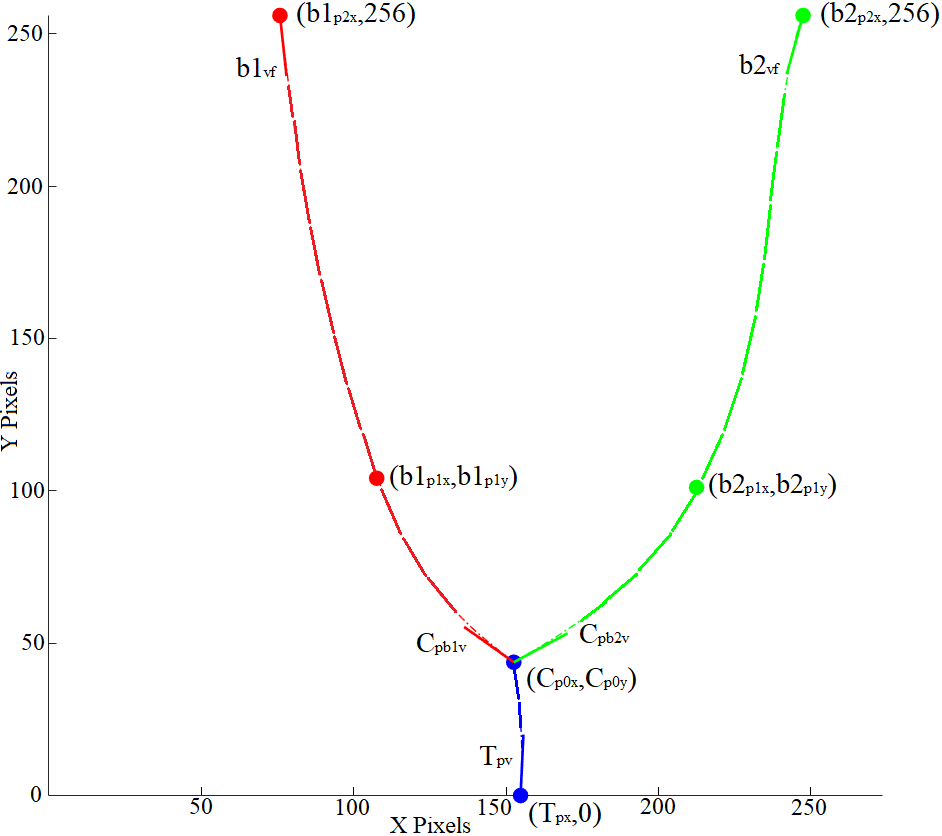}
    \caption{Visual representation of the 14 parameter Y-tree template.}
    \label{fig:YtreeSkele}
\end{figure}

The problem defined for the Genetic Algorithm was to fit a 14-parameter predefined Y tree template (Figure \ref{fig:YtreeSkele}) to a partial tree skeleton. Genetic Algorithm was selected as the optimization tool as the problem is well defined, but there are many variables to be optimized. The partial tree skeleton was found by analyzing each row of the image individually, where each blob was converted to an average blob position.

The score to be optimized was then calculated using mean absolute error (MAE) by comparing the blobs center position in the partial skeleton (Figure \ref{fig:Automated_example} (3)) to the generated Y shaped tree template in each row. An example of a fitted tree template is shown in Figure \ref{fig:Automated_example} (4). Compared to other post-processing methods such as curve fitting, a predefined skeleton can hallucinate portions of the tree past the endpoint of the partial skeleton. 

A Genetic Algorithm was set up according to the following parameters, we adopted the notation used for the parameters from~\cite{MitchellMelanie1996Aitg}, $N_p = 2000$, $T=800$, $e_{tac}=2$, $e_{tam}=2$, $P_c=0.8$ and $P_m = 0.5$.

The 14 parameters used to define the Y tree template were as follows. Firstly, to define the trunk there were 4 parameters:
\begin{itemize}
    \item $(T_{px},0)$ = starting point
    \item $(C_{p0x},C_{p0y})$ = endpoint of the trunk
    \item $T_{pv}$ = gradient of the trunk from the point $(T_{px},0)$
\end{itemize}
Using these boundary conditions, the coefficients of a quadratic curve were calculated for the trunk.

To define each branch ($i=1,2$) there were 5 parameters:
\begin{itemize}
    \item $(C_{p0x},C_{p0y})$ = starting point 
    \item $C_{pbiv}$ = gradient of the branch from $(C_{p0x},C_{p0y})$
    \item $(bi_{p1x}, bi_{p1y})$ = via point of the branch
    \item $(bi_{p2x}, 256)$ = endpoint of the branch
    \item $bi_{vf}$ = final gradient at the point $(bi_{p2x}, 256)$
\end{itemize}
Using these boundary conditions the coefficients of a two connected cubic curves were calculated for each branch. Figure \ref{fig:YtreeSkele} shows an example visual representation of the 14 parameters that define the Y tree template. 

\subsubsection{Repairing}
\begin{figure}[!htbp]
    \centering
    \includegraphics[width=0.9\textwidth]{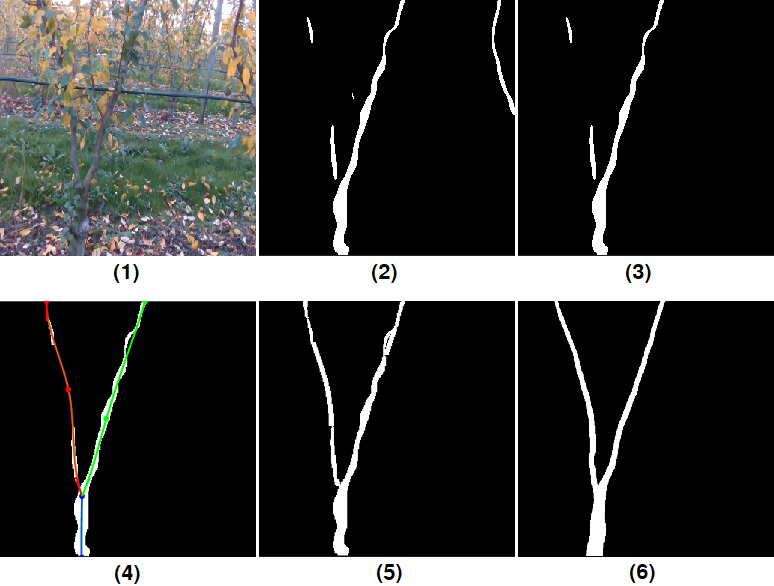}
    \caption{Example of each step in the automated process. (1) represents the RGB image; (2) and (3) are the corresponding CNN Prediction and the Filtered Prediction; (4) is the Fitted Tree Template fitted to the Filtered Prediction; (5) is the final adjusted and repaired image and (6) is the ground truth for comparison.}
    \label{fig:Automated_example}
\end{figure}

Using the Y-shaped tree template generated from the tree fitting stage, the horizontal blob thickness profile was measure along the template using the filtered prediction as the reference. Where there are gaps in the thickness profile, the missing thickness values are generated linearly. If an upper endpoint is missing, a thickness value of 4 pixels is assumed. Additionally, a minimum value of generated thickness is set as 3 pixels. These values were selected based on the average size of the upper endpoints and average minimum branch thicknesses in the initial training set. A final filter was applied to the image keeping only the largest blob, finally, the repaired image was generated (Figure \ref{fig:Automated_example} (5)). 

In this process, images were not used if more than 50\% of the branches and trunk were reconstructed or if the basic requirements of the basic Y-shaped tree filter were not met.

\subsection{Metrics} \label{Sec::Metrics}

\subsubsection{Traditional Metrics}
\begin{figure}[!htbp]
    \centering
    \includegraphics[width=0.65\textwidth]{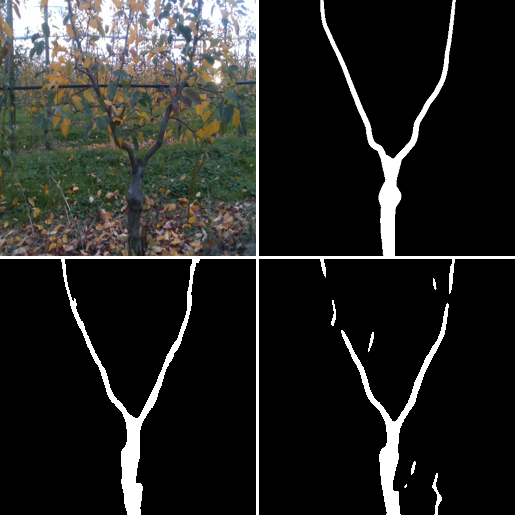}
    \caption{Example of a successful 'Good' label (bottom left) and a 'Bad' label (bottom right) with the RGB (top left) and ground truth (top right) for comparison.}
    \label{fig:IMG_GvB}
\end{figure}

To evaluate the performance of the different CNNs and generated labels, mean IOU (mIOU) and Boundary F1 (BF1) were used \citep{metrics,BF1}. The 52 images in the validation dataset were used to evaluate the CNNs. We set the distance threshold to $2$ pixels for BF1. In the application of detecting tree skeletons, where the majority of the image is the background, mIOU scored relatively close for different quality predictions of the same input image. For example, Figure \ref{fig:IMG_GvB} shows an example of a `Good' and `Bad' label, with the RGB and ground truth image for comparison. The metric values are shown in Table \ref{tab:GvB}. While BF1 score shows a large difference, mIOU shows relatively small differences in performance.

Additionally, the performance and number of successful labels were analyzed. We define a successful label as a completely connected mask with no noise, which can be used in the environment modeling process.

\subsubsection{Complete Grid Scan}
As mIOU does not accurately reflect the qualitative differences and BF1 focuses on boundary accuracy, we define an additional metric, the Complete Grid Scan (CGS), that reflects the position and thickness accuracy of blobs, and rewards good connectivity and low noise. Additionaly, CGS aims to quantitatively reflect this observed qualitative difference in predictions. 

By comparing the prediction to the ground truth, we find the error $\alpha$, as the sum of distances between centers and the difference in thickness of each prediction blob, with the closest ground truth blob measured along the horizontal rows:

\begin{equation}
    \alpha = {\displaystyle\sum_{k=1}^{n} (|p_t (k)-p_p (k)|+|t_t (k)-t_p (k)|)}
\end{equation}
where, $p_t (k)$ is the center position of the ground truth blob $k$, $p_p (k)$ is the center position of the closest predicted blob to $p_t (k)$, $t_t (k)$ is the thickness of the ground truth blob $k$ and $t_p (k)$ is the thickness of the corresponding predicted blob $k$. $n$ is the total number of comparisons.

However, this does not account for a potential difference in the number of blobs per row. Therefore, we need to consider the total number of errors, $n_e$, the difference in the number of blobs per row:

\begin{equation}
n_e = \displaystyle\sum_{j=1}^{h} |n_t (j)-n_p (j)|
\end{equation}
where $n_t (j)$ is the number of blobs in the ground truth, $n_p (j)$ is the number of blobs in the predicted image, h is the height of image in pixels. 

We consider each error in the number of blobs as the maximum possible distance error in a row, i.e. the image width in pixels, $w$. Therefore, we defined $\eta$ as the error due to incorrect number of blobs:

\begin{equation}
    \eta = w n_e
\end{equation}

We define $CGS_h$ as the combination and normalization of the two sources of error measured on the horizontal rows. 

\begin{equation}
    CGS_h = 1- \frac{\alpha + \eta}{w (n + n_e)}
\end{equation}

\begin{figure}[!htbp]
    \centering
    \includegraphics[width=0.95\textwidth]{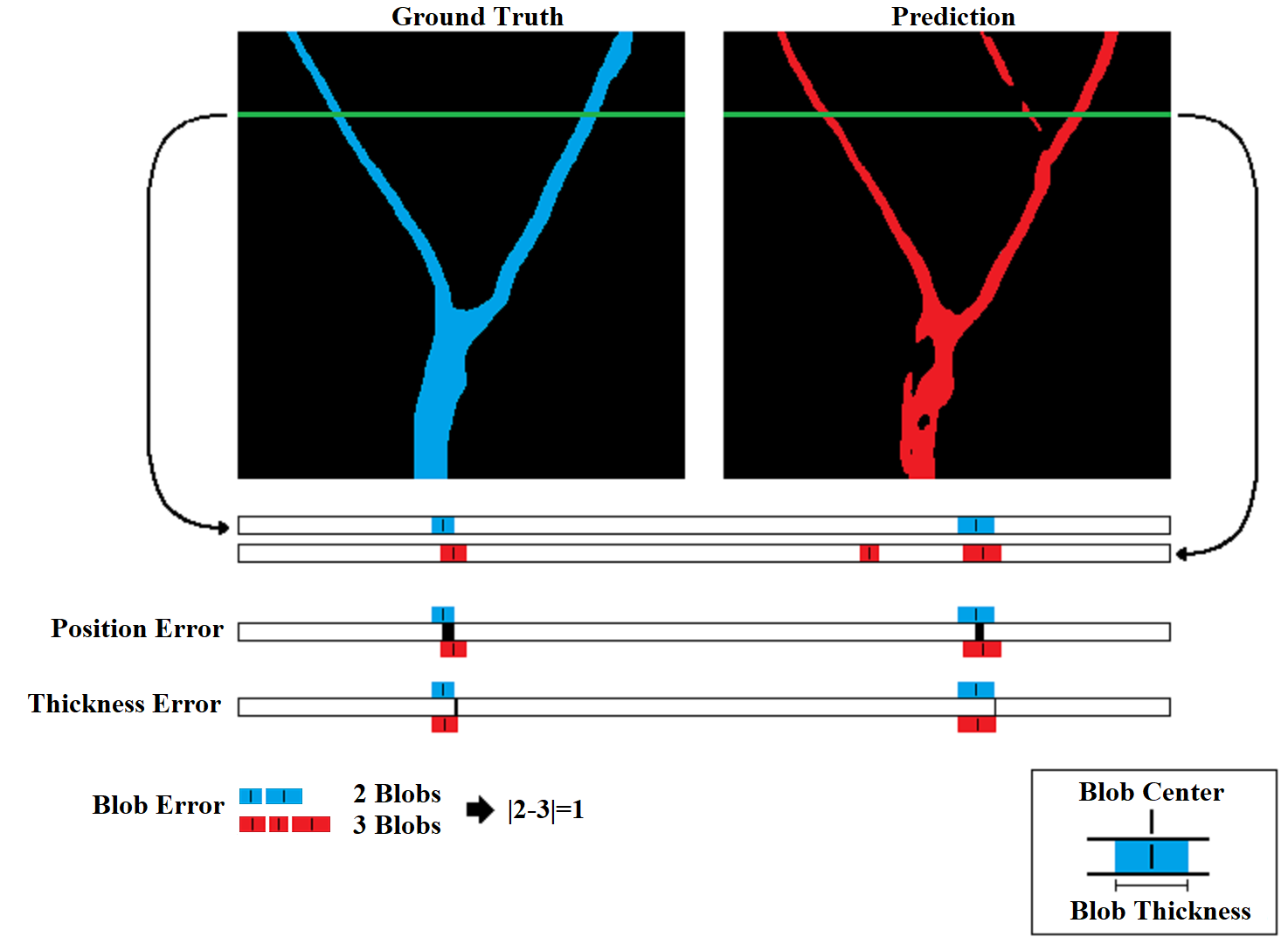}
    \caption{Example of CGS process, where a single row of a prediction is compared to the ground truth. The position and thickness of the blobs are compared. Additionally, the blob error is calculated.}
    \label{fig:CGS_example}
\end{figure}

$CGS_v$ is calculated by repeating the same process measured along the vertical columns, or, by rotating the image 90 degrees and repeating $CGS_h$. The final CGS is calculated as the average of $CGS_h$ and $CGS_v$. An example of CGS is shown in Figure \ref{fig:CGS_example} . 

The CGS score for the `Good' and `Bad' image (Figure \ref{fig:IMG_GvB}) was calculated (Table \ref{tab:GvB}). The difference in mIOU performance is small compared to BF1 and CGS, which more accurately reflects the difference in the quality of the predicted images for this application. CGS performance is an indicator of good connectivity and low noise.

\begin{table}[!htbp]
\begin{center}
    \caption{Metric evaluation on the `Good' and `Bad' predictions from Figure \ref{fig:IMG_GvB}}
    \label{tab:GvB}
    \begin{tabular}{lccc}
    \hline
    & \textbf{mIOU} & \textbf{BF1} &  \textbf{CGS}  \\ 
      \hline
      Good & 0.8028 & 0.8734 & 0.9210\\
      Bad & 0.7743 & 0.7561 & 0.7682\\
      \hline
    \end{tabular}
\end{center}
\end{table}
\vspace{-2mm}

\section{Results} \label{Results}
All methods (HITL, ATL, FBST, and CST) were evaluated on 52 images in the validation set. The quality, quantity and effort of each method were analyzed. The Manual method refers to the CNNs trained on the ground truth images. 

Two separate trials were completed, each with different initial training set sizes. In Trial 1, $|S_{T0}| = 50$ and in Trial 2, $|S_{T0}| = 20$. In each trial, the number of new images introduced each iteration was equal to the initial training set size, excluding the FBST where all unlabeled images were introduced in the first iteration.  

\subsection{Trial 1 - Initial Set Size of 50}
\begin{figure}[!htbp]
    \centering
    \includegraphics[width=1\textwidth]{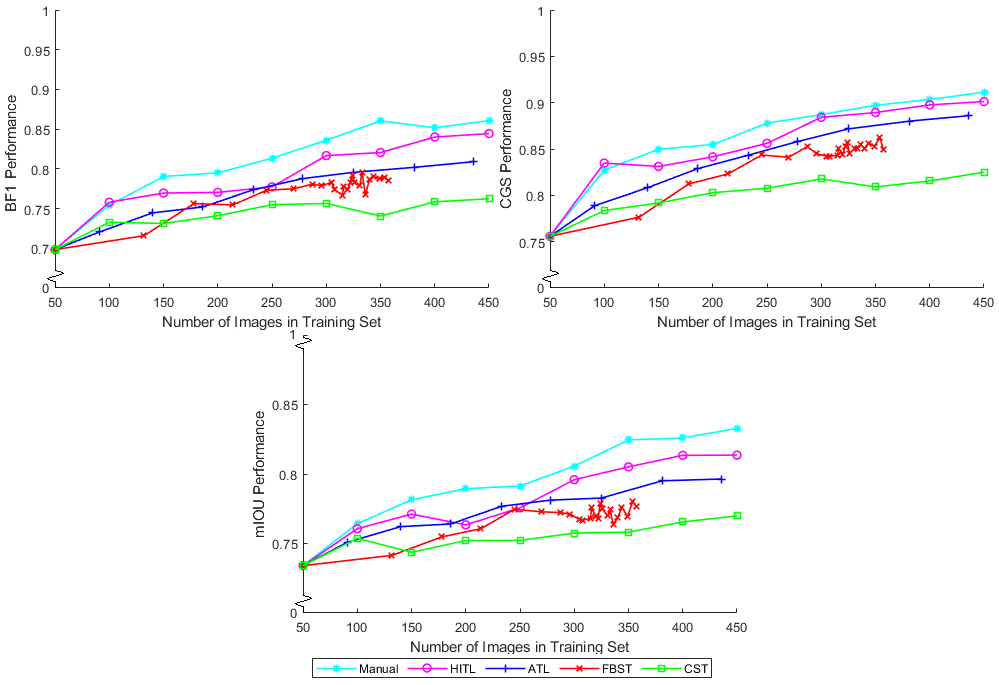}
    \caption{BF1, CGS and mIOU performance of each method during the iterative process of Trial 1.}
    \label{fig:Trial1Performance}
\end{figure}

Figure \ref{fig:Trial1Performance} shows the performance of the CNNs on the validation dataset for each method, evaluated over the iterative process. The HITL method performed relatively close to the manual method across all metrics. The ATL method achieved comparable performance in the CGS metric, scoring $2.78\%$ lower than the manual method, compared to the CST method scoring $9.47\%$ lower. ATL shows an improved performance compared to both FBST and CST over the iterative process. 

The FBST method performance showed the majority of improvements over the first four iterations ($+11.58\%$), where the process terminated after 23 iterations. Improvements slowed as iterations increased and the number of images passed through the filter reduced. Feeding the whole pool of unlabeled images allowed for $20.5\%$ of images to be approved by the filter (successfully labeled) on the first iteration. Thereby reducing the maximum number of unlabeled images that could be approved on subsequent iterations. 

For both ATL and FBST methods, not all unlabeled images were successfully labeled when the iterative process terminated. This is a result of the custom process, which acted as a final filter to the data. To compare successful labels generated in the CST process, the Y-shaped tree filter was applied to the CST generated labels, indicated by CST w/ Filter.

The performance of the successful labels for trial 1 in Tables \ref{tab:T1_IterativePerformance}.  

\begin{table}[!htbp]
    \centering
    \begin{tabular}{l|cccc}
        \textbf{Method} & \textbf{BF1} & \textbf{mIOU} & \textbf{CGS} & \textbf{Successful Labels} \\
        \hline
        HITL & 0.8996 & 0.8471 & 0.9482 & 400\\
        ATL & 0.8465 & 0.8139 & 0.9136 & 386\\
        FBST & 0.8522 & 0.8114 & 0.9077 & 307\\
        CST & 0.7989 & 0.7887 & 0.8461 & - \\
        CST w/ Filter & 0.8485 & 0.8112 & 0.9075 & 73 \\
    \end{tabular}
    \caption{Successful labeled data performance from Trial 1}
    \label{tab:T1_IterativePerformance}
\end{table}

The maximum number of unlabeled images to be labeled during the iterative process was 400. ATL successfully labeled 96.5\% of images, while FBST successfully labeled 76.8\% of images. The CST method labeled all unlabeled data, however, the performance across all metrics was lower compared to all other methods. For CST w/ Filter only 18.25\% of images were successfully labeled, with a similar performance to ATL and FBST.

\subsection{Trial 2 - Initial Set Size of 20}
\begin{figure}[!htbp]
    \centering
    \includegraphics[width=1\textwidth]{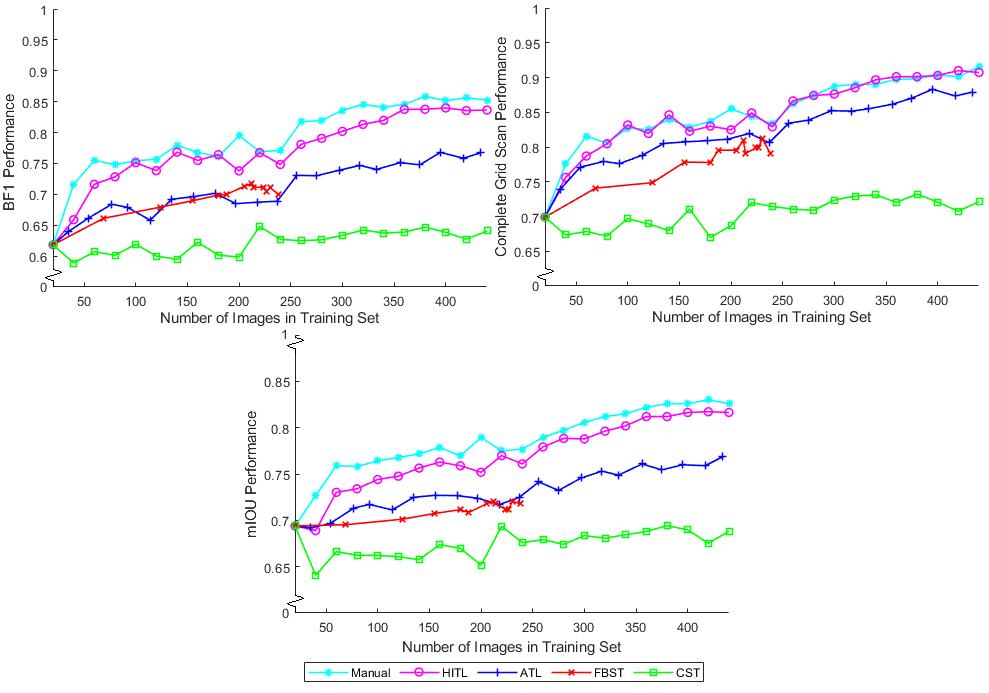}
    \caption{BF1, CGS and mIOU performance of each method during the iterative process of Trial 2.}
    \label{fig:Trial2Performance}
\end{figure}

In Trial 2, we aimed to reduce the human effort to the limit by minimizing the initial set size. Performance of all methods for each metric in Trial 2 are shown in Figure \ref{fig:Trial2Performance}. In Trial 2 the initial CNN ($CNN_0$) showed inferior performances in all metrics.

For the CST method in Trial 1 we observed good behaviors in the initial predictions compared to Trial 2, such as low noise. Over the iterations, these good behaviors were reinforced, resulting in a slow upward trend. Compared to Trial 2, where the prediction set size is smaller, and most predictions displayed bad behaviors such as increased false detections, increased noise and increased disconnections compared to Trial 1. These behaviors were reinforced over the iterations where the performance of CST remained relatively similar over the iterations.

By reducing the initial training set size, the benefits of ATL were more predominant compared to Trial 1. With lower quality initial predictions, ATL had more potential to improve adjustments and continued to trend upward over the iterations. When compared to both the CST and manual methods, ATL showed performance relatively close to the manual and HITL methods in CGS.  

With the lower quality initial prediction dataset, the FBST method struggled to pass images through the filter after a few iterations, compared to Trial 1. Indicating the FBST method is heavily dependent on the initial training dataset size, quality of the initial CNN ($CNN_0$) and total data set size ($S_{U0}$).

The performance of the successful labels for Trial 2 is shown in Table \ref{tab:T2_IterativePerformance}. ATL successfully labeled 94.1\% of images, while the FBST method successfully labeled only 49.5\% images. This shows the advantage of a higher level adjustment process.   

\begin{table}[!htbp]
    \centering
    \begin{tabular}{l|cccc}
    \textbf{Method} & \textbf{BF1} & \textbf{mIOU} & \textbf{CGS} & \textbf{Successful Labels} \\
    \hline 
    HITL & 0.8926 & 0.8441 & 0.9451 & 440\\
    ATL & 0.7984 & 0.7849 & 0.9020 & 414\\
    FBST & 0.8176 &  0.7827 & 0.8897 & 218\\
    CST & 0.7073 &  0.7339 & 0.7613 & - \\
    CST w/ Filter & 0.8223 &  0.8086 & 0.8992 & 37 \\
    \end{tabular}
    \caption{Successful labeled data performance from Trial 2}
    \label{tab:T2_IterativePerformance}
\end{table}

\begin{figure}[!htbp]
    \centering
    \includegraphics[width=1\textwidth]{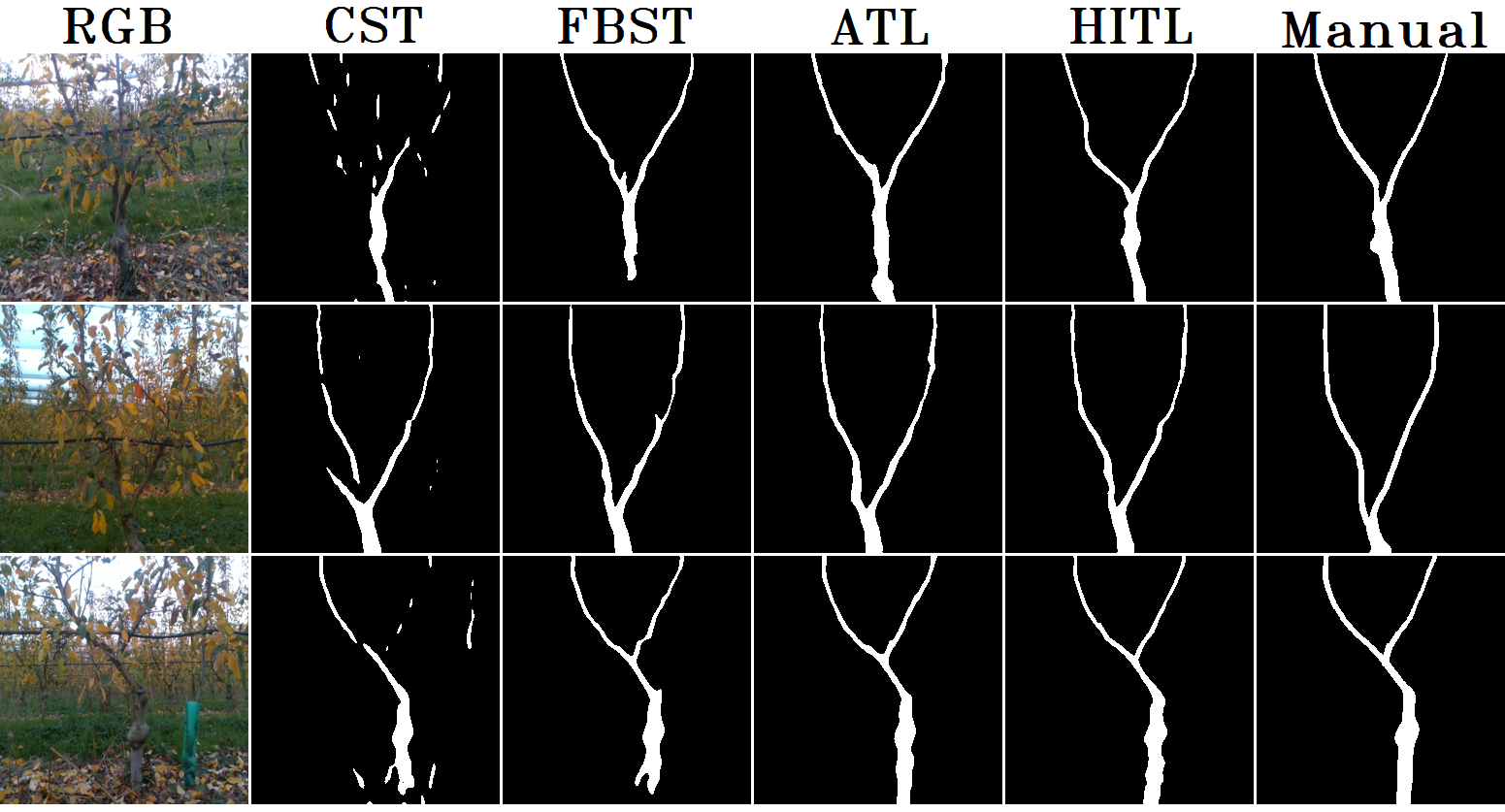}
    \caption{Examples of the labeled data from each method. The first column represents the input RGB image. The remaining columns represent the final label generated for each method, from left to right, CST, FBST, ATL, HITL and Manual respectively.}
    \label{fig:ILD}
\end{figure}

CSF labeled all unlabeled data, the performance across all metrics was lower compared to all other methods. The majority of the images produced by CST during both trials were not adequate to be used for environment modeling. This can be seen in Figure \ref{fig:ILD}. When applying the Y-shaped tree filter to labels generated in CST, only 8.81\% of images were successfully labeled.

By increasing the number of unlabeled images in the overall dataset, we would expect to observe both ATL and FBST to raise the percentage of successfully labeled images due to increased iterations. This would allow more opportunities to label unsuccessfully labeled data with improving the performance of the iterative CNNs and higher quality predictions. 

\subsection{Human Effort}

Human effort measured in minutes is summarised in Table \ref{tab:HumanEffort}. The approximate time for the HITL method (including both $S_{T0}$ and all adjusted labels) was 441.6 and 363.3 minutes for Trials 1 and 2, respectively. In Trial 2 we observed a reduction factor of 4.3 using the HITL method for this application. 

For all other iterative methods, the only effort was to create the initial training set, resulting in 175 and 70 minutes for Trial 1 and 2. We observed a reduction factor of 22.5 in manual labor for ATL, CST, and FBST methods in Trial 2. 

\begin{table}[]
    \centering
    \begin{tabular}{l|c|c}
        \textbf{Method/s} & \textbf{Trial} & \textbf{Human Effort (mins)} \\
        \hline
        Manual & - & 1575 \\
        \hline
        \multirow{2}{4em}{HITL} & 1 & 441.6 \\
         & 2 & 363.3 \\
         \hline
        \multirow{2}{8em}{ATL, FBST, CST} & 1 & 175 \\
         & 2 & 70
    \end{tabular}
    \caption{Human effort measured in minutes, for each labeling method.}
    \label{tab:HumanEffort}
\end{table}

\vspace{-2mm}

\section{Conclusion} \label{Conculsion}

In this paper, we presented an iterative training methodology for using CNNs to segment tree skeletons. To minimze human effort, we introduced Automating-the-Loop, which involves attempting to automatically replicate the adjustment made by annotators in the HITL process. We introduced a new metric, Complete Grid Scan, to indicate good connectivity and low noise in binary images. Using ATL, we successfully created a CNN with competitive performance to the manual and HITL CNNs. It was found that ATL has the greatest ratio of performance to labeling effort. For semi-supervised methods (including ATL), the quality of the custom process greatly impacted on the performance. This results in an interesting problem of balancing the custom processes to generate successful labels and minimizing human effort. Future work includes using the ATL method to label other apple tree structures by exploring other automatic processes, using these labels to accurately model the 3D farming environment, and optimization of the iterative training method. 
\clearpage

\bibliography{refs}

\end{document}